\documentclass[10pt,twocolumn,letterpaper]{article}
\usepackage{pdflscape}

\usepackage{cvpr}              
\usepackage{booktabs}
\usepackage{multirow}
\usepackage{graphicx}
\usepackage{xcolor, colortbl}
\usepackage{algorithm}
\usepackage{algorithmic}
\usepackage{amssymb}
\usepackage{amsmath}
\usepackage{booktabs} 
\usepackage{multirow} 
\usepackage{graphicx} 
\usepackage{amssymb}
\definecolor{darkgreen}{rgb}{0.0, 0.7, 0.0}
\usepackage{caption}
\usepackage{newfloat}
\usepackage{listings}
\usepackage{rotating}

\definecolor{cvprblue}{rgb}{0.21,0.49,0.74}
\usepackage[pagebackref,breaklinks,colorlinks,allcolors=cvprblue]{hyperref}

\def\confName{CVPR}
\def\confYear{2026}

\title{DrivePTS: A Progressive Learning Framework with Textual and Structural Enhancement for Driving Scene Generation}

\author{
Zhechao Wang\hspace{0.4em}Yiming Zeng \hspace{0.4em}Lufan Ma\hspace{0.4em}Zeqing Fu\hspace{0.4em}Chen Bai\hspace{0.4em}Ziyao Lin\hspace{0.4em}Cheng Lu\\[0.2em]
$^{1}$XPeng Motors
}

\begin{document}
\maketitle
\begin{abstract}
Synthesis of diverse driving scenes serves as a crucial data augmentation technique for validating the robustness and generalizability of autonomous driving systems. 
Current methods aggregate high-definition (HD) maps and 3D bounding boxes as geometric conditions in diffusion models for conditional scene generation.
However, implicit inter-condition dependency causes generation failures when control conditions change independently.
Additionally, these methods suffer from insufficient details in both semantic and structural aspects. 
Specifically, brief and view-invariant captions restrict semantic contexts, resulting in weak background modeling. 
Meanwhile, the standard denoising loss with uniform spatial weighting neglects foreground structural details, causing visual distortions and blurriness.
To address these challenges, we propose DrivePTS, which incorporates three key innovations.
Firstly, our framework adopts a progressive learning strategy to mitigate inter-dependency between geometric conditions, reinforced by an explicit mutual information constraint. 
Secondly, a Vision-Language Model is utilized to generate multi-view hierarchical descriptions across six semantic aspects, providing fine-grained textual guidance. 
Thirdly, a frequency-guided structure loss is introduced to strengthen the model's sensitivity to high-frequency elements, improving foreground structural fidelity.
Extensive experiments demonstrate that our DrivePTS achieves state-of-the-art fidelity and controllability in generating diverse driving scenes.
Notably, DrivePTS successfully generates rare scenes where prior methods fail, highlighting its strong generalization ability.
\end{abstract}
\section{Introduction}
\label{sec:intro}
Autonomous driving technology has achieved significant progress in perception tasks, such as 3D multiple-object tracking  (MOT) ~\cite{zhang2022mutr3d, pang2023standing, ding2024ada} and online local mapping (OLM)~\cite{droeschel2018efficient, li2022hdmapnet, liu2023vectormapnet}.  
Furthermore, autonomous systems based on Vision Language Action (VLA)~\cite{zhou2025opendrivevla, zhou2025autovla} demonstrate strong potential.
However, generic copilot systems without high-precision maps exhibit notable deficiencies in navigation, particularly in road selection and path planning under complex traffic scenarios.
A critical bottleneck hindering the navigation capability is the difficulty of collecting comprehensive road configurations. 
Acquiring specific traffic scenarios is both time-consuming and resource-intensive, often subject to geographical and regulatory restrictions. 
To address data scarcity, recent studies have leveraged diffusion models to synthesize diverse, controllable traffic scenes. 
Techniques such as BEVControl, MagicDrive, and others~\cite{yang2023bevcontrol, gao2023magicdrive, wen2024panacea, li2024drivingdiffusion, li2025dualdiff} integrate diverse conditional inputs, including text descriptions, camera parameters, and 3D spatial information, to generate traffic scenes that adhere to these controllable elements. 
These methods demonstrate the potential to generate diverse driving scenes for data augmentation and construct challenging scenarios during simulation validation.
\begin{figure}[t]
	\centering
	\includegraphics[width=0.465\textwidth]{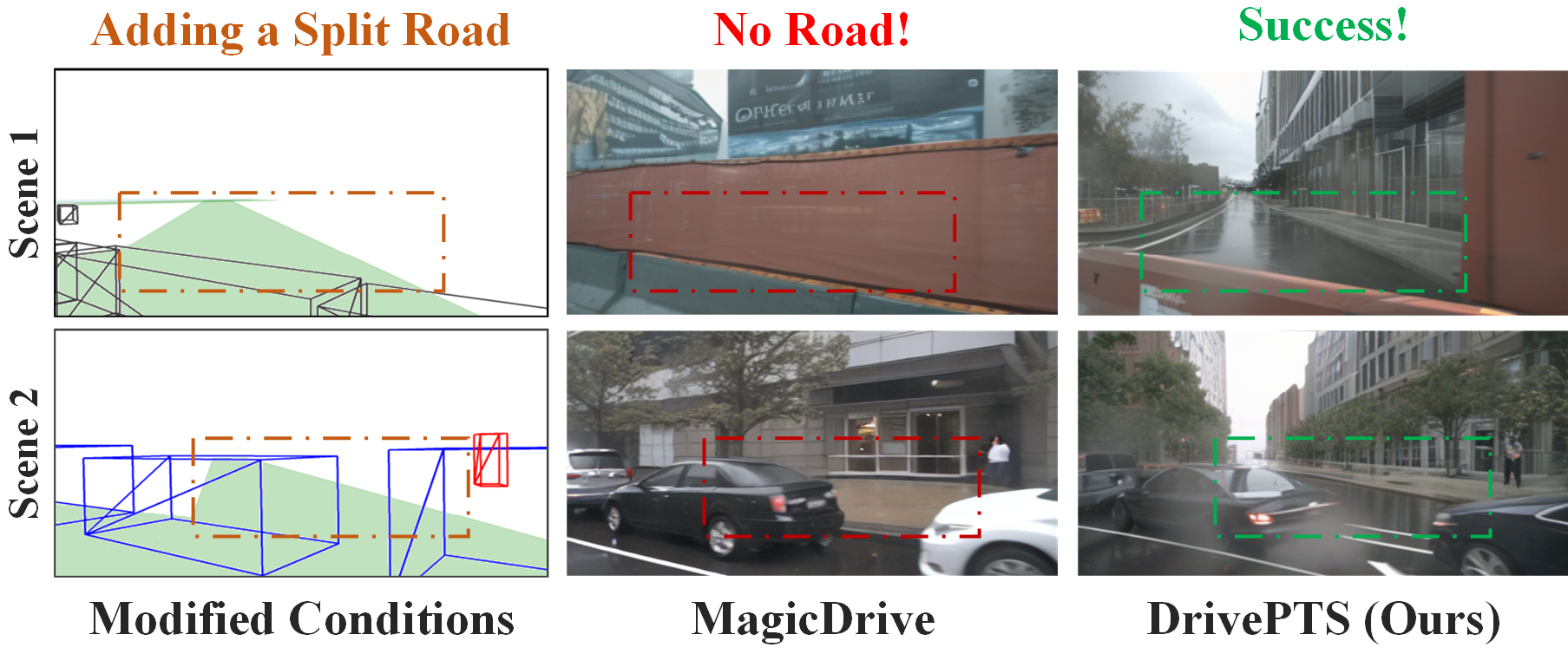}
	\caption{Comparison of various scene generation methods on the modified map layouts. 
		The first column highlights the modified regions of each map. 
		While MagicDrive fails to adapt to map modifications, our DrivePTS successfully generates scenes aligning with the updated map configurations.
	}
	\label{fig.0}
	\vspace{-14pt}
\end{figure}

Despite these advancements, existing methods exhibit significant limitations:
\textbf{1) Inter-dependency between geometric conditions.}
When 3D bounding boxes and maps are jointly learned, current approaches may suffer from overfitting to co-occurrence patterns between road structures and traffic objects. As illustrated in  Fig. \ref{fig.0}, the generator may implicitly associate a row of parked cars with a straight road or match roadblocks with the absence of roads, failing to produce corresponding changes even when map conditions are modified.
\textbf{2) Weak  textual guidance.}
Existing methods rely on brief, view-invariant textual descriptions that contain only basic information, failing to capture fine-grained and viewpoint-specific environmental details necessary for high-fidelity generation. 
These coarse-grained captions result in weak background modeling, as evidenced by higher Fréchet Inception Distance (FID) scores and qualitative visual assessment in our empirical studies.
\textbf{3) Lack of foreground details.}
Furthermore, existing methods prioritize global scene generation, treating all areas equally when computing the denoising loss, while neglecting foreground details. This results in structural distortions and blurriness in the generated objects and roads.

To address these challenges, we propose DrivePTS, a progressive learning framework with textual and structural enhancement for driving scene generation.
\textbf{1) Progressive learning strategy to reduce inter-condition dependency.} 
We propose a progressive training paradigm that decomposes the learning of geometric conditions. 
The model first learns to generate road layouts, followed by object placement and rendering. This separation ensures independent modeling of each condition, preventing dependencies between maps and bounding boxes. 
To mitigate catastrophic forgetting, the two phases are trained alternately. 
Once individual conditions are sufficiently learned, both are fed simultaneously for dual-input adaptation. 
Meanwhile, a mutual information constraint is also introduced to explicitly decouple the two geometric conditions.
\textbf{2) Fine-grained textual guidance via Vision-Language Model (VLM).}
We leverage a VLM to generate descriptions for each view across six distinct aspects: time, weather, road type, surroundings, objects, and spatial relationships. This multi-view, hierarchical textual guidance leads to more accurate and detailed scene generation.
\textbf{3) Foreground detail enhancement.}
We introduce a frequency-guided structure loss to strengthen the model's sensitivity to high-frequency structural elements, such as road edges and object contours, thereby mitigating  foreground distortions and blurriness.

Extensive quantitative and qualitative experiments are conducted on the nuScenes dataset~\cite{caesar2020nuscenes} to demonstrate the effectiveness of DrivePTS. Compared to recent state-of-the-art baselines, our method achieves significant improvements: FID decreases by 1.91, road IoU increases by 2.69, vehicle IoU increases by 0.69, and object detection NDS score improves by 1.44.
Visualization results show that our multi-view, hierarchical captions facilitate more comprehensive recovery of scene details, with clearer outlines of both objects and roads attributed to the frequency-guided structure loss. 
Additionally, DrivePTS can generate rare scene configurations, which existing methods fail to produce. This demonstrates the advantages of our progressive learning strategy in reducing inter-condition dependencies.
Our main contributions can be summarized:
\begin{itemize}
	\item We propose a novel progressive learning strategy that separates road and object learning into distinct stages, with a mutual information constraint to explicitly mitigate dependencies between geometric conditions.
	
	\item We develop a VLM-driven approach to generate multi-view hierarchical descriptions across multiple semantic dimensions, providing fine-grained textual guidance that enhances scene generation fidelity.

	\item A frequency-guided structure loss is introduced to improve structural fidelity by enhancing the model's sensitivity to high-frequency components.
\end{itemize}

\section{Related work}
\subsection{Diffusion-Based Models in Image Synthesis.}
Diffusion models achieve high-fidelity image synthesis through iterative Markov chain diffusion steps. Latent Diffusion Models (LDMs)~\cite{rombach2022high} shift this process from pixel space to compressed latent space, reducing computational overhead while maintaining image quality, fostering widespread adoption. Dhariwal~\cite{dhariwal2021diffusion} further demonstrated that diffusion models can be conditioned on text, facilitating the generation of semantically coherent and contextually relevant images. Imagic and its variants~\cite{kawar2023i, brack2024ledits++} improve control through joint optimization of images and text descriptions, enabling semantic-level manipulation. Recent findings~\cite{lian2024llmgrounded, wu2025paragraph} indicate that longer, more descriptive captions enhance generative quality. Motivated by this, we re-caption the nuScenes dataset with detailed scene descriptions to improve conditional image synthesis.

In addition, numerous diffusion-based frameworks are proposed for conditional generation with explicit spatial control through external signals.
ControlNet~\cite{zhang2023adding} augments pretrained diffusion backbones with auxiliary neural networks that integrate layout cues, segmentation maps, or depth priors to achieve spatially aligned synthesis.
GLIGEN~\cite{li2023gligen} introduces a gated self-attention mechanism to ground image generation on bounding boxes while maintaining consistency with textual prompts.
In contrast, T2I-Adapter~\cite{mou2024t2i} employs lightweight, modular adapters that inject conditioning signals into frozen diffusion models via element-wise operations.
Its simplicity and efficiency make T2I-Adapter particularly suitable for our decoupled learning scheme based on geometric layout conditions.
\subsection{Controllable Editing Driving Scenes.}
The growing demand for large-scale annotated driving-scene datasets, coupled with high manual labelling costs, necessitates efficient synthetic data generation methods~\cite{gao2023magicdrive, zhou2024simgen, hu2024drivingworld, hu2024drivingworld}. 
LDM-based approaches~\cite{lu2024wovogen, swerdlow2024street, wang2024drivedreamer, zhang2024perldiff,  swerdlow2024street, li2025dualdiff} generate diverse driving scene views by conditioning on geometric and textual guidance, ensuring geometric consistency and multi-view alignment.
Recently, subject-driven customization has gained traction in driving scene generation. Inspired by Subject Diffusion~\cite{ma2024subject}, SubjectDrive~\cite{huang2025subjectdrive} introduces a Subject Prompt Adapter, along with a subject bank to enable flexible target replacement for improved sampling diversity.
DriveEditor~\cite{liang2025driveeditor} offers a unified framework for object repositioning, insertion, deletion, and replacement through a learned reconstruction task.
SceneCrafter~\cite{zhu2025scenecrafter} enables the insertion or removal of foreground objects and the modification of global attributes, such as weather and time of day.
MVPbev~\cite{liu2024mvpbev} extends these capabilities with camera-pose editing, supporting generalizable viewpoint changes across different driving camera settings.
However, these methods overlook map editing, limiting the diversity of road topologies presented in synthetic datasets. Our work seeks to fill this gap.

\begin{figure*}[h] 
	\centering  
	\includegraphics[width=0.98\textwidth]{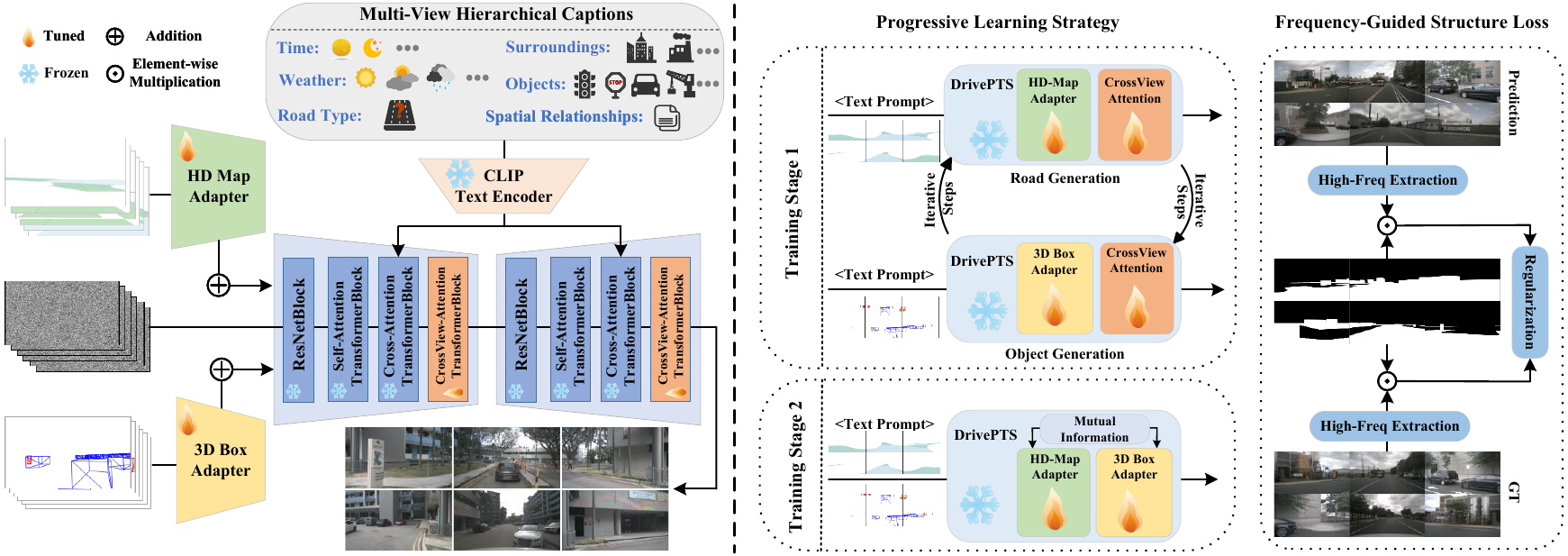} 
	\caption{The overall architecture of our proposed DrivePTS. The left part illustrates our adopted generative network structure, while the center depicts the training process corresponding to the proposed progressive learning strategy. The right part highlights the implementation of frequency-guided structure loss.}  
	\label{fig.1}  
\end{figure*}  
\section{Methods}
In this section, we first provide preliminary background on the denoising process of Stable Diffusion and the T2I-Adapter architecture for geometric conditioning. 
We then present our DrivePTS framework, as illustrated in Fig.~\ref{fig.1}, including: \textbf{1)} a progressive learning strategy to mitigate inter-dependencies between geometric conditions, \textbf{2)} VLM-driven multi-view hierarchical scene description generation to enrich textual guidance, \textbf{3)} frequency-guided structure loss for improving structural details, and \textbf{4)} an overall training objective that unifies these components.
\subsection{Preliminary}
\subsubsection{Stable Diffusion.} We implement our method based on Stable Diffusion (SD)~\cite{rombach2022high}, a text-to-image LDM that performs denoising in the latent space through a UNet denoiser. The training objective is formulated as:
\begin{equation}
	L_{\text{diff}} = \mathbb{E}_{Z_{t},C,\epsilon,t}\left[\left\|\epsilon - \epsilon_{\theta}(Z_t,C)\right\|_2^2\right]
\end{equation}
where $Z_{t}$
represents the noised latent feature at timestep $t$. 
$C$ denotes condition information  and $\epsilon_{\theta}$ is the UNet denoiser function.
During inference, the latents $Z_t$ are gradually denoised through $\epsilon_\theta$ conditioned on $C$ and $t$. Finally, the denoised latent is converted to an image by the decoder of the autoencoder.
\subsubsection{T2I-Adapter.} 
Previous approaches typically incorporate various conditions via ControlNet~\cite{zhang2023adding}, which involves intricate branched structures. Since our framework is designed to process two geometric conditions separately, utilizing parallel composed ControlNets would significantly increase computational cost.
We therefore adopt the lightweight T2I-Adapter~\cite{mou2024t2i}. Specifically, we treat geometric conditions as image inputs and feed them into corresponding adapters, extracting multi-scale features that are directly injected into the UNet encoder. The process of extracting condition features is formulated as:
\begin{align}
	F_c = T(C), \quad \hat{F}_{enc}^{i} = F_{enc}^{i} + F_{c}^{i},\ i\in \{1,2,3,4\},
\end{align}
where $C$ denotes the geometric condition input, $T$ is the T2I-Adapter network, $F_c$ are the extracted condition features, $F_{enc}^{i}$ represents the encoder features at the $i$-th downsampling stage, and condition features are additively integrated at each scale.

\subsection{Progressive Learning for Geometric Conditions}
For driving scene generation, geometric constraints can be broadly classified into two types: HD maps and 3D bounding boxes. 
This paper proposes a progressive learning strategy, emphasizing the principle that the generative model should first focus on constructing the road before introducing traffic objects. 
To achieve this principle, the two types of geometric conditions are processed separately and incorporated into the model through distinct training stages. 

\textbf{Stage 1: Separated Learning Geometric Conditions.} 

\textit{{Road Generation:}}
HD maps and text descriptions are utilized as condition inputs to generate the road and background elements. 
During this process, regions occupied by traffic objects are explicitly excluded from consideration, allowing the model to focus exclusively on learning the distinctive features of road infrastructure and environmental context. 
This targeted approach ensures that fundamental scene components are properly established before introducing dynamic elements.

\textit{{Object Generation:}}
Following the road-learning process, the model transitions to object generation, where 3D bounding boxes and textual descriptions serve as conditional inputs to produce traffic-related objects. 
During this process, regions outside the scope of traffic-related objects are intentionally disregarded, which makes the model concentrate on accurately placing and rendering these objects.

To mitigate the catastrophic forgetting of road-generation capabilities during object generation learning, we employ an alternating training strategy. This strategy ensures that both road and object generation are effectively learned with minimal mutual dependency.

\textbf{Stage 2: Dual Geometric Conditions Adaptation.}

During this stage, both HD map and 3D bounding box conditions are simultaneously fed into the model. The cross-view module remains frozen while both adapters are fine-tuned. 
This setup enables effective adaptation to concurrent inputs while mitigating inter-condition dependency that may arise during cross-view interactions.
Additionally, a mutual information (MI) constraint is adopted to reduce inter-condition dependency between these two geometric features, encouraging each condition branch to focus independently on its respective semantic content.
The MI constraint is implemented with a modified InfoNCE loss~\cite{oord2018representation}:
\begin{equation}
	{L}_{\text{MI}} = \mathbb{E}_{(f_m, f_b^+)} \left[ \log \frac{\exp(\text{sim}(f_m, f_b^+))}{\sum\limits_{j=1}^{N} \exp(\text{sim}(f_m, f_{b,j}))} \right],
\end{equation}where $f_m$ and $f_b^+$ represent map and corresponding box features, $f_{b,j}$ includes all box features, $\text{sim}(\cdot, \cdot)$ a similarity measure, and $N$ is the total number of samples. 
By minimizing this loss, the model is encouraged to reduce the similarity between the map and its corresponding box features, thereby facilitating independent learning for each condition.

\subsection{Multi-View Hierarchical Scene Description Generation}
Previous studies utilize the concise and view-invariant scene descriptions provided by the existing dataset as textual conditions for image generation. However, these brief descriptions often lack the details required to reconstruct complex environmental contexts and fail to capture the inherent heterogeneity across varied perspectives. 
Several studies~\cite{nie2024compositional,lian2024llmgrounded, wu2025paragraph} demonstrate that fine-grained descriptions substantially enhance the quality and controllability of generation, facilitating precise manipulation of scene components. 
These findings highlight the potential of leveraging detailed textual inputs to improve the fidelity and diversity of driving scene generation.
Nevertheless, manually crafting such fine-grained descriptions is labor-intensive and time-consuming. 

To address this challenge, we employ vision-language models (VLMs) to automatically generate high-quality and detailed scene descriptions.
We fine-tune an open-source multimodal VLM with LoRA~\cite{hu2022lora} on driving-domain datasets (DriveLM~\cite{sima2024drivelm} and nuscenes-OmniDrive~\cite{wang2025omnidrive}) to reduce hallucinations and enhance scene understanding.
To further align generated captions with factual scene attributes, we apply direct preference optimization (DPO)~\cite{rafailov2023direct} as a reinforcement correction mechanism, guiding the model toward human-consistent outputs. During description generation, the VLM processes all viewpoint images simultaneously to ensure cross-perspective consistency and semantic completeness.
For each viewpoint, we extract a structured representation encompassing six semantically distinct aspects:
\begin{itemize}
	\item \textbf{Time}: Specifies the time of day (e.g., daytime, night, dawn, dusk), which influences scene lighting and ambiance.
	\item \textbf{Weather}: Describes atmospheric conditions that impact visibility and the environment (e.g., sunny, overcast, foggy, rainy, snowy, etc.).
	\item \textbf{Road Type}: Details the layout of the road, such as straight road, left-turn lane, T-junction, roundabout, etc.
	\item \textbf{Surroundings}: Identifies the scene's location, such as commercial areas, rural areas, residential streets, construction zones, etc.
	\item \textbf{Objects}: Lists both stationary and moving objects in the scene, including cars, pedestrians, traffic signs, trees, bicycles, cones, buildings, etc.
	\item \textbf{Spatial Relationships}: Describes the geometric relationships and interactions among objects in the scene.
\end{itemize}

By explicitly modeling these elements, our crafted descriptions provide a comprehensive textual input that captures both global scene characteristics and fine-grained object relationships.

\subsection{Frequency-Guided Structure Loss}
Existing scene generation methods treat all regions equally when computing the denoising loss, failing to emphasize critical foreground details such as boundaries and textures. However, effective scene generation should not merely involve coarse filling of regions with plausible content. It also requires accurate restoration of edges and textures, which correspond to high-frequency components.

To address this limitation, we introduce a frequency-guided structure loss emphasizing high-frequency details in foregrounds, including the regions of roads and objects. This enhancement improves structural fidelity and visual clarity of the generated images.
Specifically, we utilize Fourier transforms~\cite{brigham1988fast} to extract high-frequency components through a high-pass filter, formulated as:

\begin{equation}
	\begin{aligned}
		\mathcal{H}(x) &= \mathcal{F}^{-1}(M(\omega) \cdot \mathcal{F}(x)), \\
		M(\omega) &= 
		\begin{cases}
			0, & \|\omega\| \le \tau, \\
			1, & \|\omega\| > \tau,
		\end{cases}
	\end{aligned}
\end{equation}
where $\mathcal{F}$ and $\mathcal{F}^{-1}$ denote the Fourier transform and its inverse, respectively. $M(\omega)$ represents the high-pass filter that selectively retains frequency components above the threshold $\tau$. Besides, $\omega$ refers to the frequency coordinates, $\|\omega\|$ denotes the frequency magnitude, and $\tau \in (0,1)$ determines the cutoff frequency for high-frequency retention, empirically set to 0.5.
The frequency-guided structure loss is then formulated as:

\begin{equation}
	L_{\text{freq}} = \left\| \mathcal{H}(x_{\text{pred}}) - \mathcal{H}(x_{\text{target}}) \right\|_2^2.
\end{equation}

\subsection{Overall Training Objective}
The overall training objective integrates multiple loss components to achieve distinct goals within our progressive learning strategy. 

\textbf{Stage 1:} 
During this stage, road and object generation are learned separately with an alternating training strategy.

The loss function for \textit{road generation}  is defined as:
\begin{equation}
	L_{\text{road}} = L_{\text{diff}} \odot (M_{\text{map}} + M_{\text{bg}}) + \lambda_{\text{freq}} \cdot L_{\text{freq}} \odot M_{\text{map}},
\end{equation}
where $M_{\text{map}}$ and $M_{\text{bg}}$ denote binary masks for map and background regions, respectively, $\odot$ represents element-wise multiplication for region-specific loss, and $\lambda_{\text{freq}}$ is the weighting coefficient for frequency-guided structure loss.

The loss function for \textit{object generation}  is given by:
\begin{equation}
	L_{\text{object}} = L_{\text{diff}} \odot M_{\text{box}} + \lambda_{\text{freq}} \cdot L_{\text{freq}} \odot M_{\text{box}},
\end{equation}
where $M_{\text{box}}$ denotes the binary mask of bounding box regions.

\textbf{Stage 2:}
Building upon the localized optimization achieved in Stage 1, this stage progresses toward modeling both maps and objects simultaneously:
\begin{equation}
	L_{\text{stage2}} = L_{\text{diff}} + \lambda_{\text{freq}} \cdot L_{\text{freq}} \odot (M_{\text{map}} + M_{\text{box}}) + \lambda_{\text{MI}} \cdot L_{\text{MI}},
\end{equation}
where $\lambda_{\text{MI}}$ is the weighting coefficient for mutual information constraint.
\section{Experiments}
\subsection{Setup}
\subsubsection{Dataset.} 
We evaluate our method on the nuScenes dataset, which contains 1,000 examples of 360° street-view scenes captured by six cameras. Following the official configuration, we utilize 700 scenes (28,130 frames) for training and 150 scenes (6,019 frames) for validation. 
For objects, we focus on ten categories: car, bus, truck, trailer, motorcycle, bicycle, construction vehicle, pedestrian, barrier, and traffic cone. Each category is represented by distinctly colored bounding boxes, which serve as inputs for the box adapter. 
Regarding maps, we concentrate on three primary types: driving area, lane, and pedestrian crossing. These layouts are also color-coded and processed by the map adapter.
\subsubsection{Evaluation Metrics.} 
Following prior works, we evaluate the proposed method based on metrics that measure  generation fidelity and controllable accuracy of objects and roads. 
Specifically, Frechet Inception Distance (FID) is adopted to measure the realism of synthesized images.
For controllability, we test the generated validation set with pre-trained perception models, CVT~\cite{zhou2022cross} and BEVFusion~\cite{liu2022bevfusion}. 
Metrics such as nuScenes Detection Score (NDS), mean Average Precision (mAP), and mean Average Orientation Error (mAOE) are used to quantify alignment with ground truth annotations.
\subsubsection{Implementation Details.}
Our framework builds upon Stable Diffusion 2.1, initializing the UNet with pre-trained weights. 
Training consists of two stages: geometric conditions are learned separately for 60k steps, followed by dual condition adaptation for 10k steps. 
Trade-off weights are set as $\lambda_\text{freq} = 0.5$ for frequency-guided structure loss and $\lambda_\text{MI} = 0.05$ for mutual information constraint.
We use the AdamW optimizer~\cite{loshchilov2017decoupled} with a learning rate of $6 \times 10^{-5}$. During inference, the DDIM sampler~\cite{songdenoising} is configured with 25 steps, and the classifier-free guidance (CFG) scale is set to 3. The resolution of the generated images is $224 \times 480$ pixels.
For multi-view scene descriptions, we leverage Qwen2.5-VL-72B~\cite{bai2025qwen2}, achieving state-of-the-art performance in visual understanding and textual generation. Additional implementation details are provided in the supplementary material.

\begin{table}[t]
	\centering
	\caption{
		Comparison of generation fidelity and controllability across different driving-view generation methods on the NuScenes {validation} set. Panacea* denotes our reproduced results based on the officially provided generated dataset.  Outcomes demonstrating superior performance are highlighted in \textbf{bold}. $\uparrow$ / $\downarrow$ indicates that a higher/lower value is better. }
	\label{table:1}
	\resizebox{\linewidth}{!}{
		\begin{tabular}{@{}lcc|ccc|cc@{}}
			\toprule
			\multirow{2}{*}{Method} & \multirow{2}{*}{\begin{tabular}[c]{@{}c@{}}Synthesis\\ resolution\end{tabular}} & \multirow{2}{*}{FID$\downarrow$} & \multirow{2}{*}{mAP$\uparrow$} & \multirow{2}{*}{NDS$\uparrow$} & \multirow{2}{*}{mAOE$\downarrow$} & Road & Vehicle\\
			&&&&&&mIoU$\uparrow$&mIoU$\uparrow$\\
			\midrule
			Oracle & -- & -- & 35.54 & 41.20 & 0.56 & 73.68 & 34.80  \\
			\midrule
			BEVGen  & --  & 25.54 & -- & -- & -- & 50.20 & 5.89 \\
			BEVControl & 224 $\times$ 400 & 24.85 & -- & -- & -- &  60.80 & 26.80 \\
			MagicDrive & 224 $\times$ 480 & 16.20 & 12.30 & 23.32 & -- &  61.05 & 27.01 \\
			Panacea* & 256 $\times$ 512 & 16.96  & 11.65 & 22.40 & 0.82 & 57.11 & 22.77\\
			PerLDiff & 256 $\times$ 704 & 13.36  & {15.24} & {24.05} & {0.78} & {61.26} & {27.13}\\
			\rowcolor{gray!25} Ours & 224 $\times$ 480 & \textbf{11.45}  & \textbf{15.37} & \textbf{25.49} & \textbf{0.76} & \textbf{63.95} & \textbf{27.82}\\
			\bottomrule
		\end{tabular}
		
	}
\end{table}
\subsection{Main Results}
\subsubsection{Realism Evaluation.}
In the realm of image synthesis, lower FID scores indicate better reconstruction quality of real driving scene styles. As shown in Tab. \ref{table:1}, our method achieves an FID score of 11.45, representing the highest realism among all tested approaches.
Our approach significantly outperforms existing solutions, including MagicDrive and Panacea, and even surpasses the recent state-of-the-art method Perldiff by approximately 16.7\%. This substantial improvement demonstrates that fine-grained hierarchical descriptions are highly effective for scene reconstruction.
\subsubsection{Geometrical Controllability.}
Our method shows remarkable performance in segmentation tasks, achieving a road mIoU of 63.95 and surpassing the second-best method by 2.69 points, as shown in the last two columns of Table~\ref{table:1}. This can be attributed to our progressive learning strategy, which prioritizes learning road layouts prior to objects. Additionally, there is a modest improvement in vehicle segmentation.
In terms of 3D object detection, our approach outperforms previous methods by attaining the highest mAP and NDS scores, indicating strong geometric alignment with ground truth annotations. The reductions in orientation and positioning errors can be attributed to the combined benefits of our frequency-guided structure loss and progressive learning strategy.
\subsubsection{Training Support for BEV Segmentation.}
Generative models have been widely recognized as powerful tools for data augmentation, significantly enhancing the generalization capabilities of perception models. 
To evaluate this potential, we leverage synthesized datasets to improve segmentation performance on the NuScenes test set. 
Since the test dataset lacks annotations for objects except roads, we focus our evaluation on improving road segmentation performance.
The results presented in the second row of Tab.~\ref {Tab.augmentation-35k} demonstrate that augmenting training data with optimally annotated real data (i.e., combining the NuScenes training set with the real validation set) yields substantial performance improvements. 
Furthermore, our synthetic validation set achieves competitive augmentation results that closely approximate the performance gains obtained with real validation data, while significantly outperforming other driving scene generation approaches.
\begin{table}[H]
	\centering
	\caption{
		Performance comparison for the boosting performance of BEV segmentation models using synthesized dataset on the NuScenes \textit{test} set using CVT. 
		The “\textit{train} + Real \textit{val}” configuration serves as a benchmark, representing the ideal upper performance limit achievable. 
		The numbers in parentheses indicate the performance disparity relative to the “\textit{train} + Real \textit{val}” configuration.
	}
	\label{Tab.augmentation-35k}
	\resizebox{0.65\linewidth}{!}{
		\begin{tabular}{@{}l|c@{}}
			\toprule
			Training & Road mIoU$\uparrow$ \\
			\midrule
			\textit{train} & 65.83 \\
			\textit{train} + Real \textit{val} & 67.53 \\
			\textit{train} + Syn. \textit{val} (MagicDrive) & 66.12 {(-1.41)} \\
			\textit{train} + Syn. \textit{val} (Panacea) & 66.60 {(-0.93)} \\
			\textit{train} + Syn. \textit{val} (PerLDiff) & 65.74 {(-1.79)} \\
			\rowcolor{gray!25}\textit{train} + Syn. \textit{val} (Ours) & 67.49 {(-0.04)} \\
			\bottomrule
		\end{tabular}
	}
\end{table}
\subsection{Ablation Study}
\subsubsection{Effectiveness of Component Designs.}
To validate the contribution of each component, we conduct comprehensive ablation studies, as shown in Tab.~\ref{table:ablation}.
We evaluate three key components: Multi-View Hierarchical Descriptions (MHD), Frequency-Guided Structure Loss (FGSL), and Mutual Information Constraint (MIC).
Without any of these components, the vanilla model demonstrates poor generation quality.
In particular, its performance in geometric control is noticeably inferior to prior methods such as MagicDrive.
This indicates that the observed performance improvements stem from the proposed components rather than the T2I-Adapter-based architecture.
Introducing MHD alone significantly reduces the FID to 12.03, while improving Road mIoU to 61.22 and Vehicle mIoU to 26.49. This indicates that richer textual inputs provide more detailed semantic guidance, thereby enhancing both image realism and scene controllability.
Applying FGSL independently also yields substantial improvements, with FID reduced to 14.47, Road mIoU increased to 62.92, and Vehicle mIoU reaching 26.95. The gains in controllability metrics, particularly over MHD alone, indicate that the emphasis on high-frequency details effectively refines the structural generation of roads and objects.
Combining MHD and FGSL leads to further performance gains, reducing FID to 11.68 while improving both segmentation metrics. 
Finally, incorporating MIC helps the model better adapt to the joint conditioning of HD maps and 3D bounding boxes, achieving the best overall performance.
\begin{table}[t]
	\centering
	\caption{
		Ablation study on the effectiveness of Multi-View Hierarchical Descriptions (MHD), Frequency-Guided Structure Loss (FGSL), and Mutual Information Constraint (MIC) in driving scene generation.
	}
	\label{table:ablation}
	\footnotesize 
	\resizebox{0.9\linewidth}{!}{ 
		\begin{tabular}{@{}ccc|c|cc@{}}
			\toprule
			MHD & FGSL & MIC & FID$\downarrow$ & Road mIoU$\uparrow$ & Vehicle mIoU$\uparrow$ \\
			\midrule
			-- & -- & -- & 15.10 & 59.77 & 25.80 \\
			\checkmark & -- & -- & 12.03 & 61.22 & 26.49 \\
			-- & \checkmark & -- & 14.47 & 62.92 & 26.95 \\
			\checkmark & \checkmark & -- & 11.68 & 63.60 & 27.16 \\
			\rowcolor{gray!25}
			\checkmark & \checkmark & \checkmark & \textbf{11.45} & \textbf{63.95} & \textbf{27.82} \\
			\bottomrule
	\end{tabular}}
\end{table}
\subsubsection{Proper Setting of Iterative Steps.}
\begin{figure}[t]
	\centering
	\includegraphics[width=0.45\textwidth]{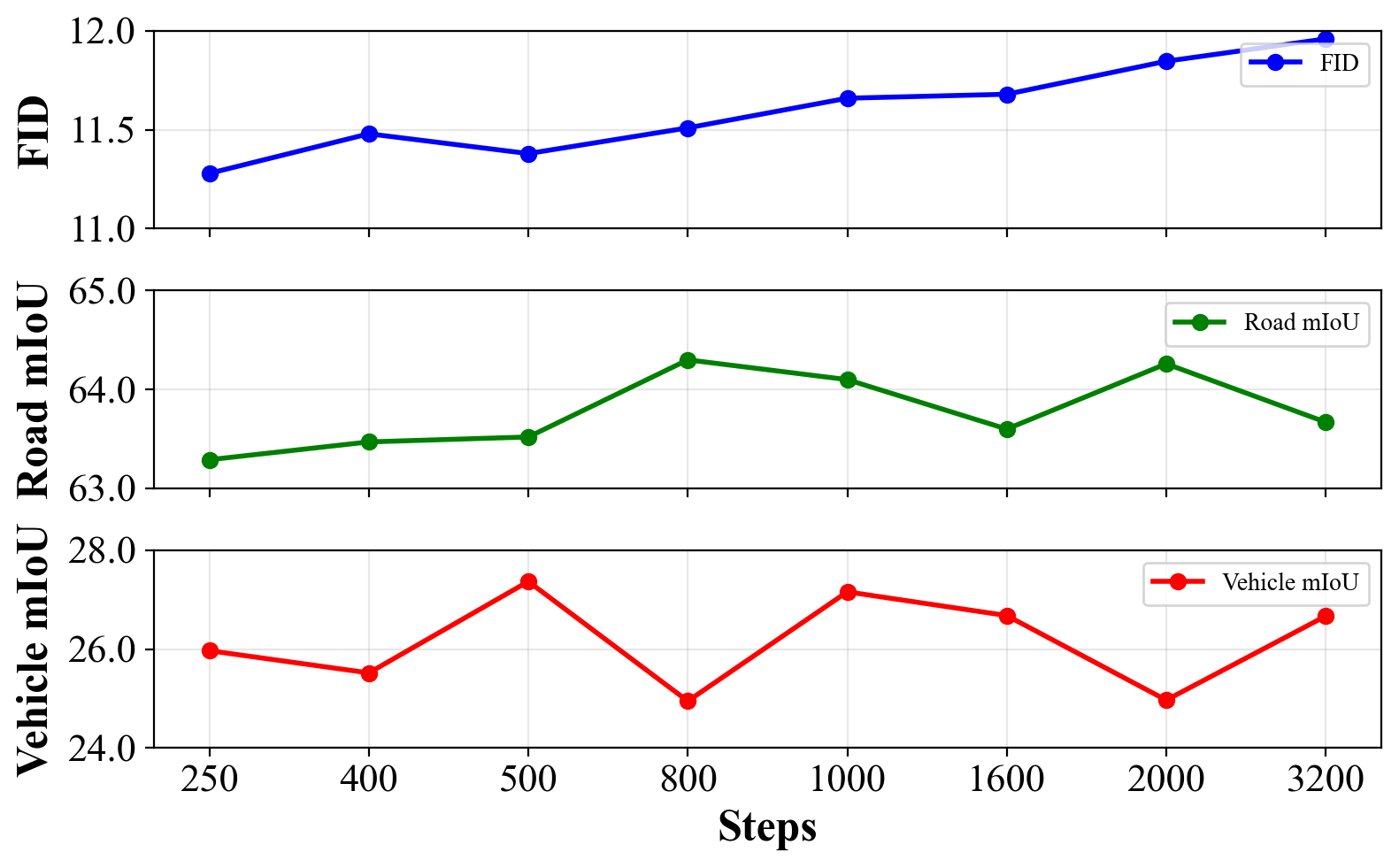}
	\caption{Impact of different iterative steps on the FID and geometry controllability of generated images.}
	\label{fig.5}
	\vspace{-8pt}
\end{figure}

During Stage 1 of our progressive learning strategy, alternating training between map and bounding box conditions requires proper selection of the iterative step interval. Fig.~\ref{fig.5} analyzes this critical hyperparameter across different step configurations.
The experimental results show that step variations impact geometric controllability more than FID scores, with the latter remaining within an acceptable fluctuation range.
The step size determines how long the network learns each condition before switching to the other. 
Shorter steps cause premature switching before the network adequately learns the current condition, thereby degrading performance.
Conversely, excessively long steps may cause catastrophic forgetting of previous conditions.
Furthermore, the condition learned at the termination of Stage 1 also significantly influences final generation quality, as recently acquired knowledge exhibits stronger retention. 
When Stage 1 concludes at 60k iterations, configurations with 400, 800, and 2000 steps terminate during road learning, achieving superior road controllability. Configurations with 250, 500, 1000, 1600, and 3200 steps conclude during object learning, yielding more accurate vehicle positioning and shapes.
We set the iterative step interval to 1000 to balance road and object generation quality, providing an optimal trade-off between the two geometric conditions.
\begin{table}[t]
	\centering
	\caption{Ablation study on coefficients $\lambda_{\text{freq}}$ and $\lambda_{\text{MI}}$. The ablation on $\lambda_{\text{freq}}$ is conducted in Stage 1, and the optimal $\lambda_{\text{freq}}$  is then applied to stage 2 for the $\lambda_{\text{MI}}$ ablation.}
	\label{table:lambda}
	\footnotesize
	
	\begin{minipage}{0.49\textwidth}
		\begin{minipage}[t][][b]{0.47\textwidth}
			\centering
			\renewcommand{\arraystretch}{0.982}  
			\resizebox{0.95\textwidth}{!}{
				\begin{tabular}{@{}c|cc@{}}
					\toprule
					$\lambda_{\text{freq}}$ & Road mIoU$\uparrow$ & Vehicle mIoU$\uparrow$ \\
					\midrule
					0 & 61.22 & 25.49 \\
					0.25 & 63.47 & 26.31 \\
					\cellcolor{gray!25}	0.5 & \cellcolor{gray!25}\textbf{63.60} & \cellcolor{gray!25}\textbf{27.16} \\
					0.75 & 63.01 & 26.84 \\
					1 & 61.95 & 26.17 \\
					\bottomrule
				\end{tabular}
			}%
		\end{minipage}%
		\begin{minipage}[t][][b]{0.47\textwidth}
			\centering
			\resizebox{0.95\textwidth}{!}{
				\begin{tabular}{@{}c|cc@{}}
					\toprule
					$\lambda_{\text{MI}}$ & Road mIoU$\uparrow$ & Vehicle mIoU$\uparrow$ \\
					\midrule
					0 & 63.60 & 27.16 \\
					\cellcolor{gray!25}	0.05 & \cellcolor{gray!25}\textbf{63.95} & \cellcolor{gray!25}\textbf{27.82} \\
					0.1 & 63.61 & 27.75 \\
					0.15 & 63.12 & 27.51 \\
					0.2 & 62.65 & {26.93} \\
					\bottomrule
				\end{tabular}
			}%
		\end{minipage}
	\end{minipage}
\end{table}
\subsubsection{Hyperparameter Analysis for Loss Coefficients.}
An ablation study is conducted on the coefficients of frequency-guided structure loss and mutual information constraint, as shown in Tab.~\ref{table:lambda}.
Increasing the frequency-guided structure loss weight $\lambda_{\text{freq}}$  notably improves road and vehicle generation accuracy by preserving fine-grained structural details. However, excessively high weights suppress overall generation quality due to over-emphasis on local details at the expense of global coherence. To strike a balance, we set $\lambda_{\text{freq}}$ to 0.5.
For the mutual information constraint, which reduces feature similarity between box and map representations during feature extraction, higher $\lambda_{\text{MI}}$ values excessively disrupt the relationships between these modalities. Since meaningful spatial correlations naturally exist between objects and road layouts in driving scenarios, our objective is to reduce inter-condition dependency while preserving essential geometric relationships. To this end, $\lambda_{\text{MI}}$ is set to 0.05 to maintain optimal performance.

\subsection{Visualisation Results}
\begin{figure}[h]
	\centering
	\includegraphics[width=0.45\textwidth]{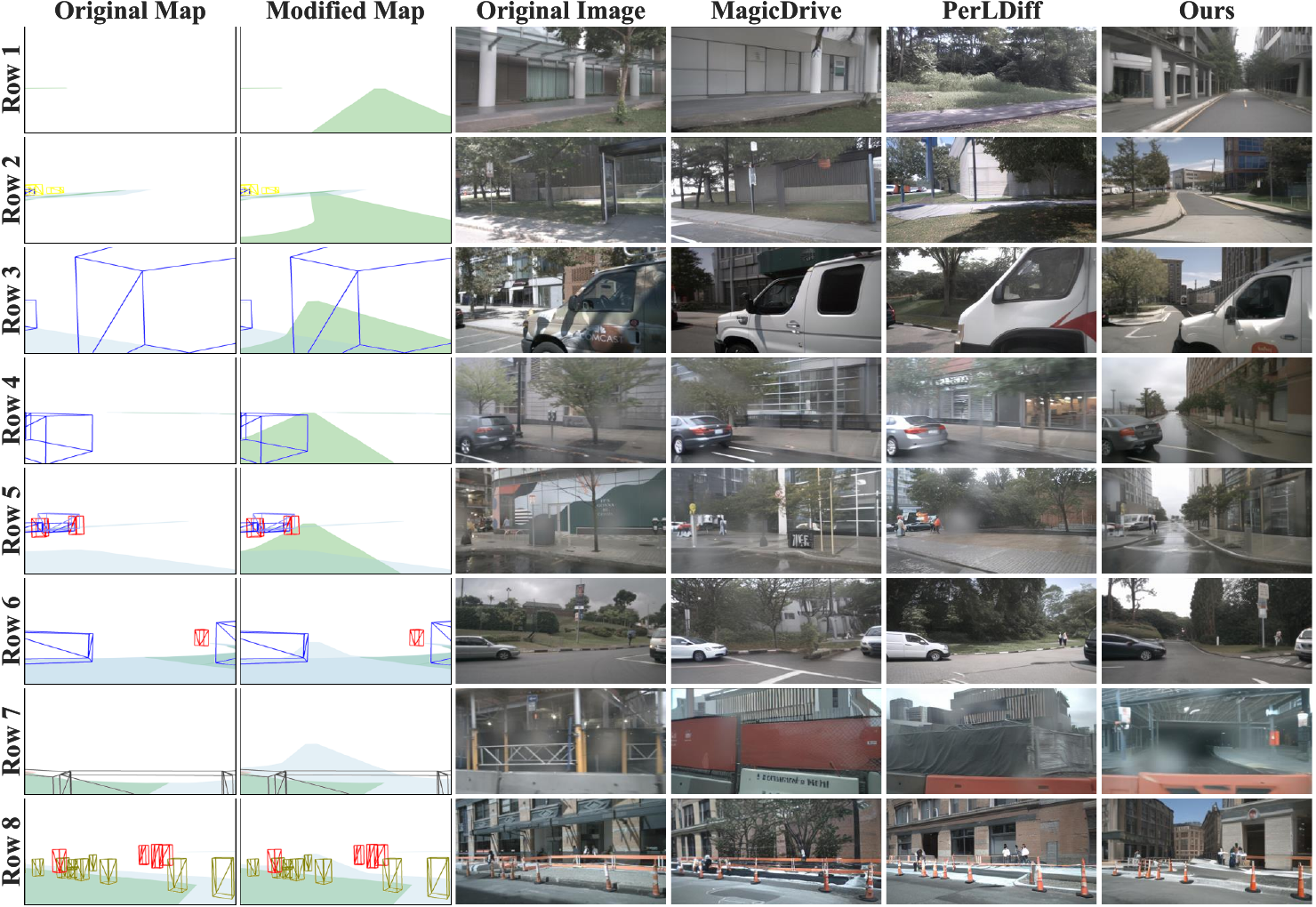}
	\caption{Examples of controllable road layout generation via HD map editing. Each row shows a case where the map is modified to introduce new road structures, with the generated image accurately reflecting these changes.}
	\label{fig.add_split_compare_big}
		\vspace{-5pt}
\end{figure}
\begin{figure}[h]
	\centering
	\includegraphics[width=0.48\textwidth]{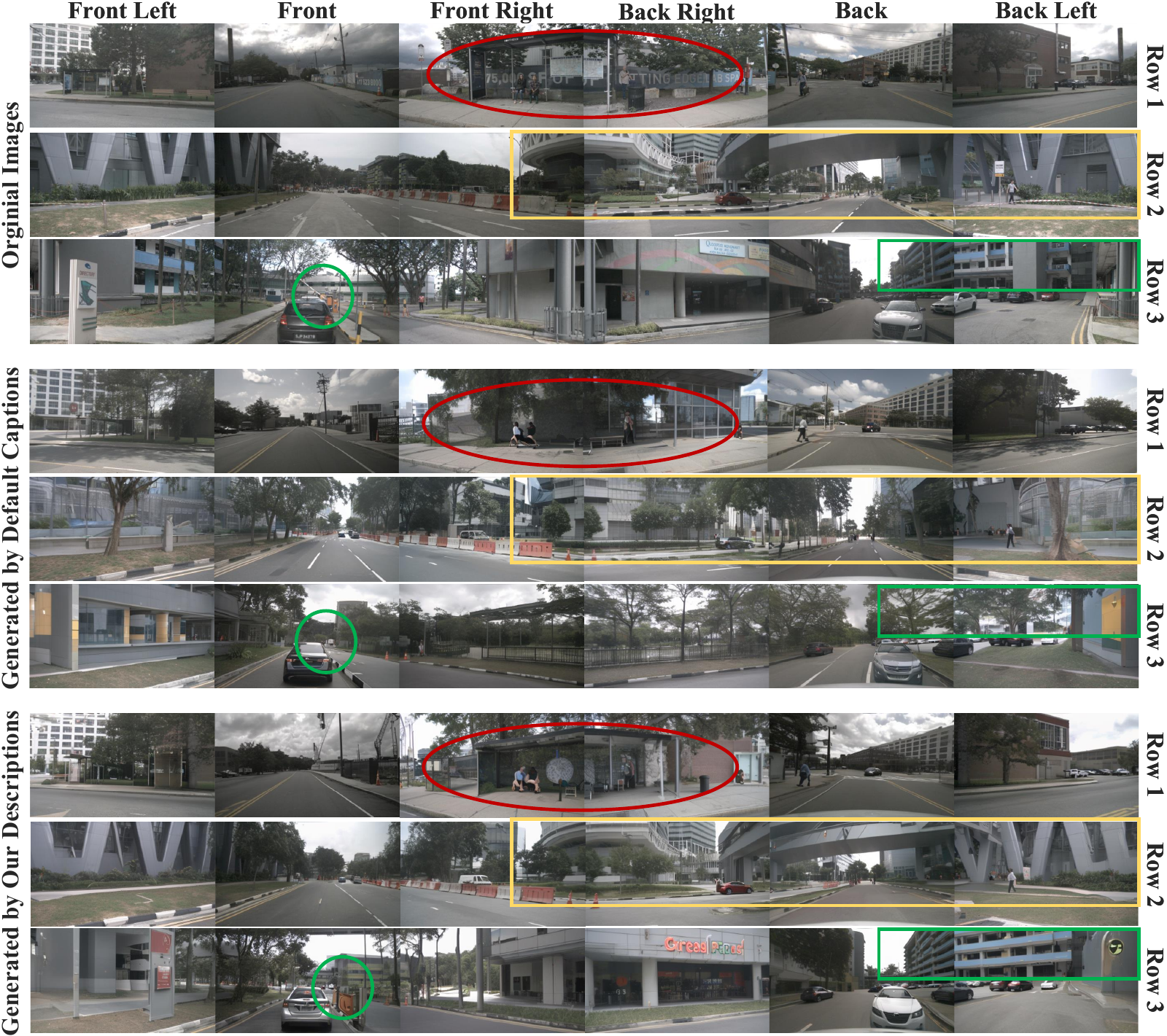}
	\caption{Visualization of the impact of textual descriptions on scene reconstruction. Rows with the same index denote the same scene. Improvements brought by multi-view hierarchical  descriptions are highlighted with various colored circles or rectangles.}
	\label{fig.vis-various-caption}
		\vspace{-8pt}
\end{figure}
\begin{figure}[t]
	\centering
	\includegraphics[width=0.48\textwidth]{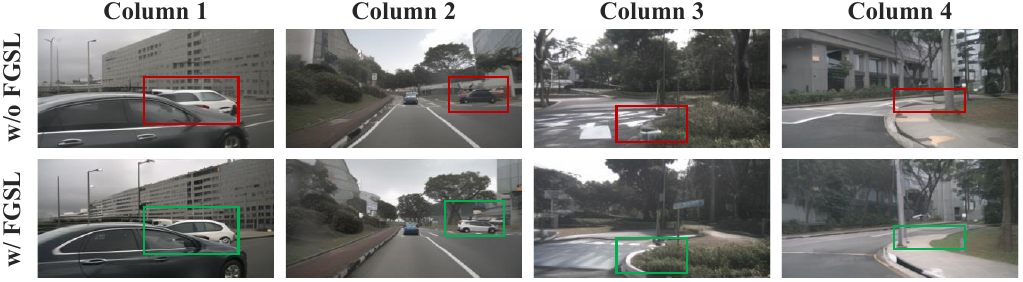}
	\caption{Qualitative comparison of scene generation with and without frequency-guided structure loss. Columns 1-2 show vehicle structure improvements, while columns 3-4 show road boundary enhancements. Red boxes highlight structural distortions without the proposed loss, whereas green boxes indicate successful mitigation with the loss.
	}
	\label{fig.with-spatial-enhancement-loss}
		\vspace{-5pt}
\end{figure}
\subsubsection{Controllable Road Layout Generation.}
Fig.~\ref{fig.add_split_compare_big} demonstrates controllable road layout generation through map editing across diverse scenarios.
Rows 1--2 show the transformation of static architectural regions into branch roads, showcasing the model's ability to synthesize plausible road structures in previously non-navigable areas.
Rows 3--4 present side road creation adjacent to parked vehicles, a challenging scenario where previous methods overfittingly link roadside parking with straight roads.
Rows 5--6 demonstrate intersection modifications, converting a T-junction into a four-way intersection by adding a road.
Rows 7--8 reveal the model's capacity to generate coherent road segments behind barriers such as fences and construction blocks, scenarios absent from existing datasets.
These results highlight the model's strong generalization capabilities and potential for creative scene synthesis. Extended examples are provided in the appendix, including cases of specific road deletion.
\subsubsection{Scene Reconstruction via Multi-view Hierarchical Descriptions.}
To evaluate the effect of textual details on scene generation, we compare images synthesized using default captions with those generated using our multi-view hierarchical descriptions. 
As shown in Fig.~\ref{fig.vis-various-caption}, our detailed descriptions lead to a more accurate reconstruction of key scene elements.
In Row 1, the red circle highlights a successfully recovered bus stop.
Row 2 shows a faithful reconstruction of a modern building and skybridge (yellow rectangle). 
In Row 3, a parking barrier gate (green circle) and blue multi-story residential building (green rectangle) are clearly restored. 
Moreover, when vehicle colors are explicitly mentioned, consistent results are generated, such as the red car in Row 2's back-right view and the white car in Row 3's back view.
These qualitative improvements demonstrate that our multi-view hierarchical descriptions significantly enhance the reconstruction of infrastructure, object attributes, and architectural details, which are elements often missed with default captions.
The superior visual quality observed in these examples corroborates our quantitative FID improvements.
\subsubsection{Visual Structure Improvement via Frequency-Guided Structure Loss. }
As demonstrated in Fig.~\ref{fig.with-spatial-enhancement-loss}, the incorporation of frequency-guided structure loss substantially improves the structural integrity and visual quality of generated scenes. 
Column 1 illustrates that the partially occluded white vehicle suffers from geometric distortions without this loss, while its shape is well-preserved when applying the frequency guidance. 
Column 2 reveals challenges in distant vehicle generation, where insufficient geometric cues lead to incomplete structures (red box), which are effectively restored through our proposed loss (green box).
Columns 3 and 4 examine structural improvements in road generation. The lack of high-frequency emphasis results in blurred and misaligned road boundaries (red box). 
Conversely, our frequency-guided structure loss recovers sharp contours and ensures better adherence to geometric layouts.

\section{Conclusion}
This paper presents DrivePTS, a controllable driving scene generation framework. 
Its progressive learning strategy facilitates the diffusion model to learn geometric conditions while reducing inter-condition dependencies. 
With the assistance of VLMs, multi-view hierarchical descriptions are automatically generated to enhance both realism and diversity of scene generation. 
Furthermore, we propose a frequency-guided structure loss that focuses on generation details by improving the model's sensitivity to high-frequency components, thereby effectively mitigating structural distortions and blurriness.
Extensive quantitative and qualitative experiments demonstrate the effectiveness of each component in our framework and showcase its superior generation capabilities compared to existing methods. 
Notably, DrivePTS can synthesize rare scene samples absent from existing datasets, a hard task where existing methods fail, highlighting its strong generalization ability.

{
    \small
    \bibliographystyle{IEEEtran}
    \bibliography{aaai2026}
}

\clearpage
\setcounter{page}{1}
\maketitlesupplementary

\section{Appendix}
\subsection{Details on Multi-View Hierarchical Description Generation}
In previous works on driving scene generation, text-based conditional inputs often lack the multi-view and fine-grained descriptions necessary for high-fidelity and contextually consistent reconstruction. 
To overcome this limitation, we leverage the advanced multimodal model Qwen2.5-VL-72B-Instruct to generate comprehensive descriptions across six distinct semantic aspects for each view. 
We fine-tune VLM with LoRA~\cite{hu2022lora} on driving-domain datasets (DriveLM~\cite{sima2024drivelm} and nuscenes-OmniDrive~\cite{wang2025omnidrive}) to reduce hallucinations and enhance scene understanding.
To further align generated captions with factual scene attributes, we apply direct preference optimization (DPO)~\cite{rafailov2023direct} as a reinforcement correction mechanism, guiding the model toward human-consistent outputs.
This facilitates more precise control over the generation process and improves realism and diversity in the synthesized scenes.
Furthermore, to accommodate the constraints of the CLIP text encoder within the Stable Diffusion UNet framework, the total token count for each scene description is constrained to a maximum of 77 tokens. 
The specific prompt employed for the VLM is presented in Listing~\ref{lst:prompt}.

\begin{figure*}[t]
	\centering
	\includegraphics[width=0.9\textwidth]{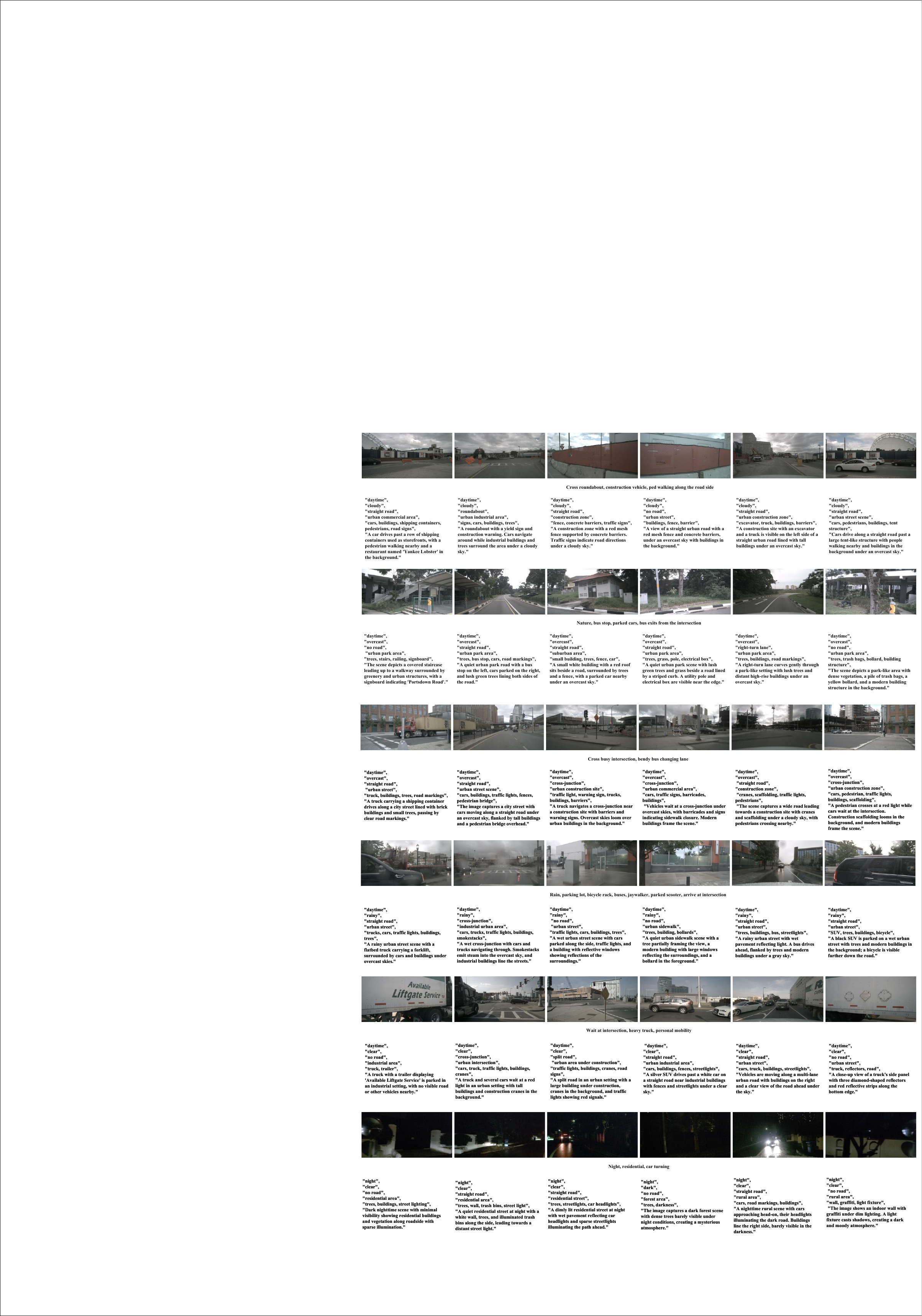}
	\caption{Comparison of original dataset captions and VLM-driven multi-view hierarchical descriptions. 
		Each group shows: (top) six-view images, (middle) original dataset captions, and (bottom) our multi-view hierarchical descriptions.
	}
	\label{fig.caption}
\end{figure*}

As shown in Fig.~\ref{fig.caption}, there are some example scenes paired with their multi-view hierarchical descriptions.
Beyond the fundamental temporal information, the generated texts effectively convey weather conditions, including overcast, rainy, clear sky, etc. 
Additionally, it categorizes various road types, such as straight roads, cross-junctions, right-turn lanes, roundabouts, split roads, and scenarios with no road. 
Furthermore, the descriptions identify the types of objects and infrastructure present in the scene and articulate their spatial relationships.

\subsection{More Qualitative Results}
\begin{figure*}[t]
	\centering
	\includegraphics[width=0.85\textwidth]{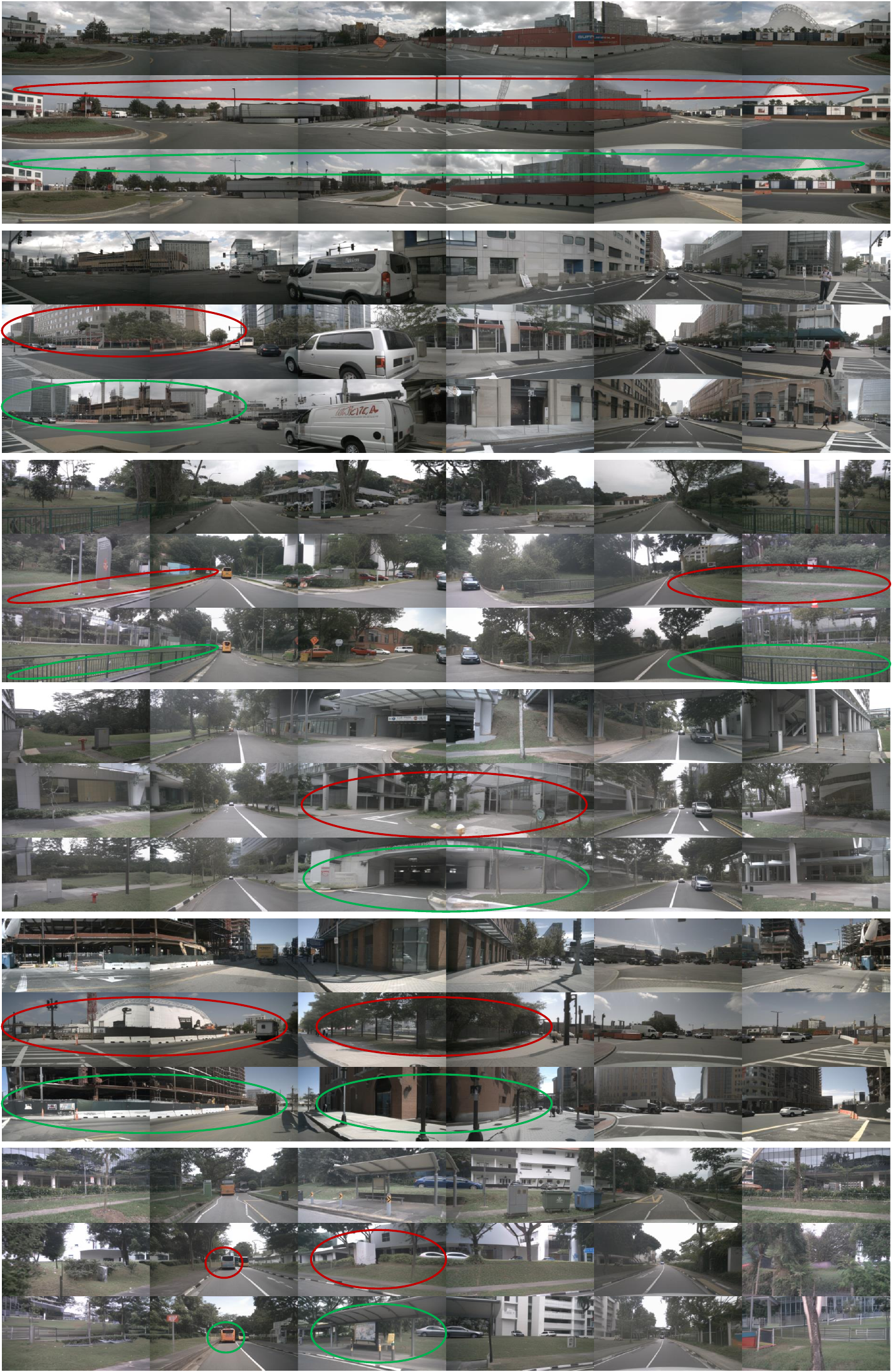}
	\caption{Qualitative comparison of scene reconstruction quality between original dataset captions and our multi-view hierarchical descriptions.
		For each example: original image (top), reconstruction from original captions (middle), and our results (bottom).
		Red regions indicate missing details with original captions, while green regions highlight successful recovery through our fine-grained descriptions.}
	\label{fig.des_comparable_1}
\end{figure*}

\begin{figure*}[t]
	\centering
	\includegraphics[width=0.85\textwidth]{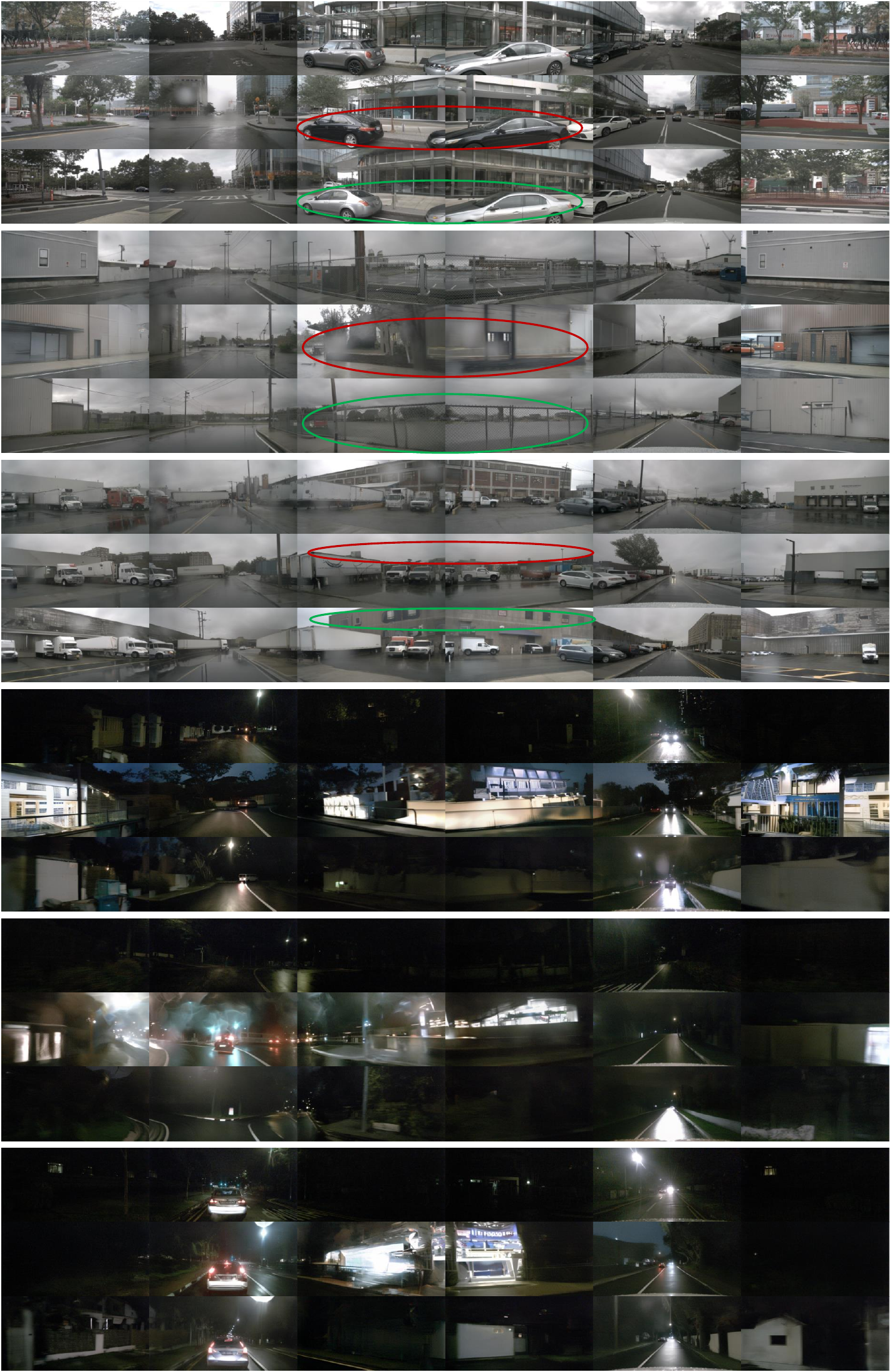}
	\caption{Qualitative comparison of scene reconstruction quality between original dataset captions and our multi-view hierarchical descriptions.
		Notably, in night scene generation, original captions tend to produce hallucinated illuminated areas due to insufficient contextual information, while our multi-view hierarchical descriptions faithfully reconstruct authentic nighttime atmospheres.}
	\label{fig.des_comparable_2}
\end{figure*}

\begin{figure*}[t]
	\centering
	\includegraphics[width=0.9\textwidth]{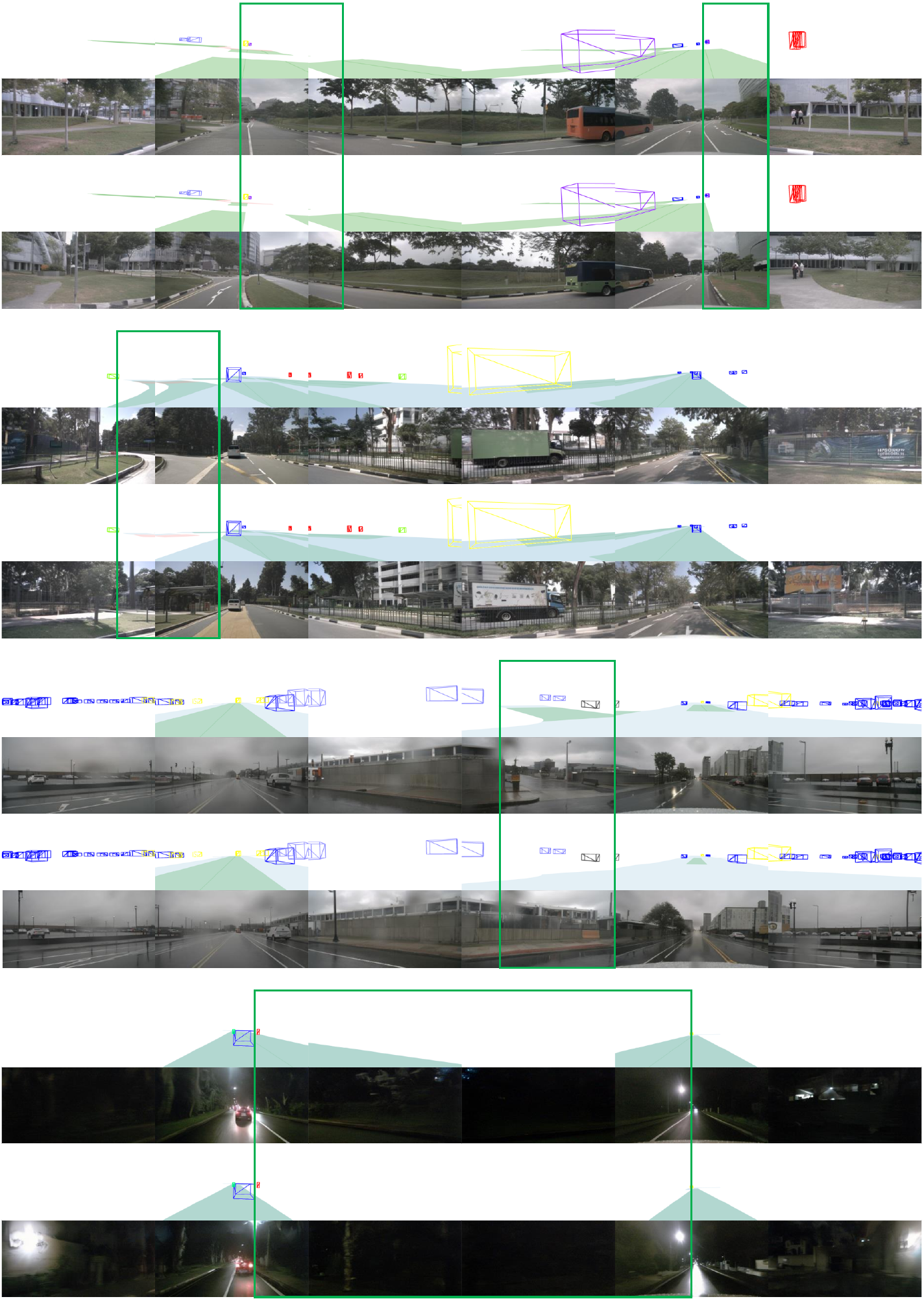}
	\caption{Examples of targeted road removal in driving scene generation using our DrivePTS framework. Each group shows: (1) original geometric conditions, (2) original scene generation, (3) modified geometric conditions after road removal, and (4) updated scene generation. Green boxes indicate areas corresponding to the geometric modifications.}
	\label{fig.appendix-del-road}
\end{figure*}

\begin{figure*}[t]
	\centering
	\includegraphics[width=0.9\textwidth]{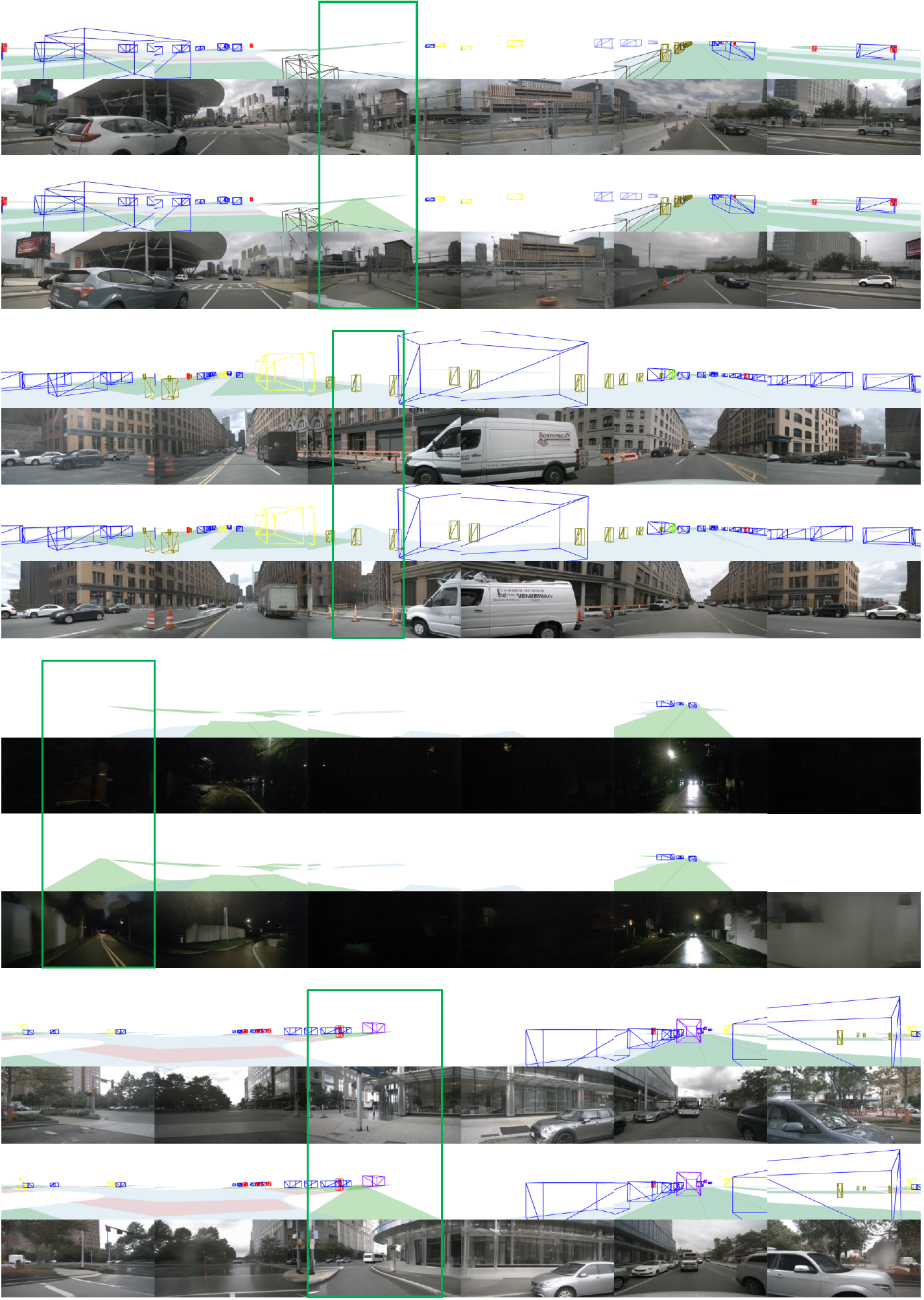}
	\caption{Examples of targeted road addition in driving scene generation using our DrivePTS framework. The visualization follows the same four-row format as above, where the third row shows geometric conditions with added roads.}
	\label{fig.appendix-add-split-road-1}
\end{figure*}

\begin{figure*}[t]
	\centering
	\includegraphics[width=0.9\textwidth]{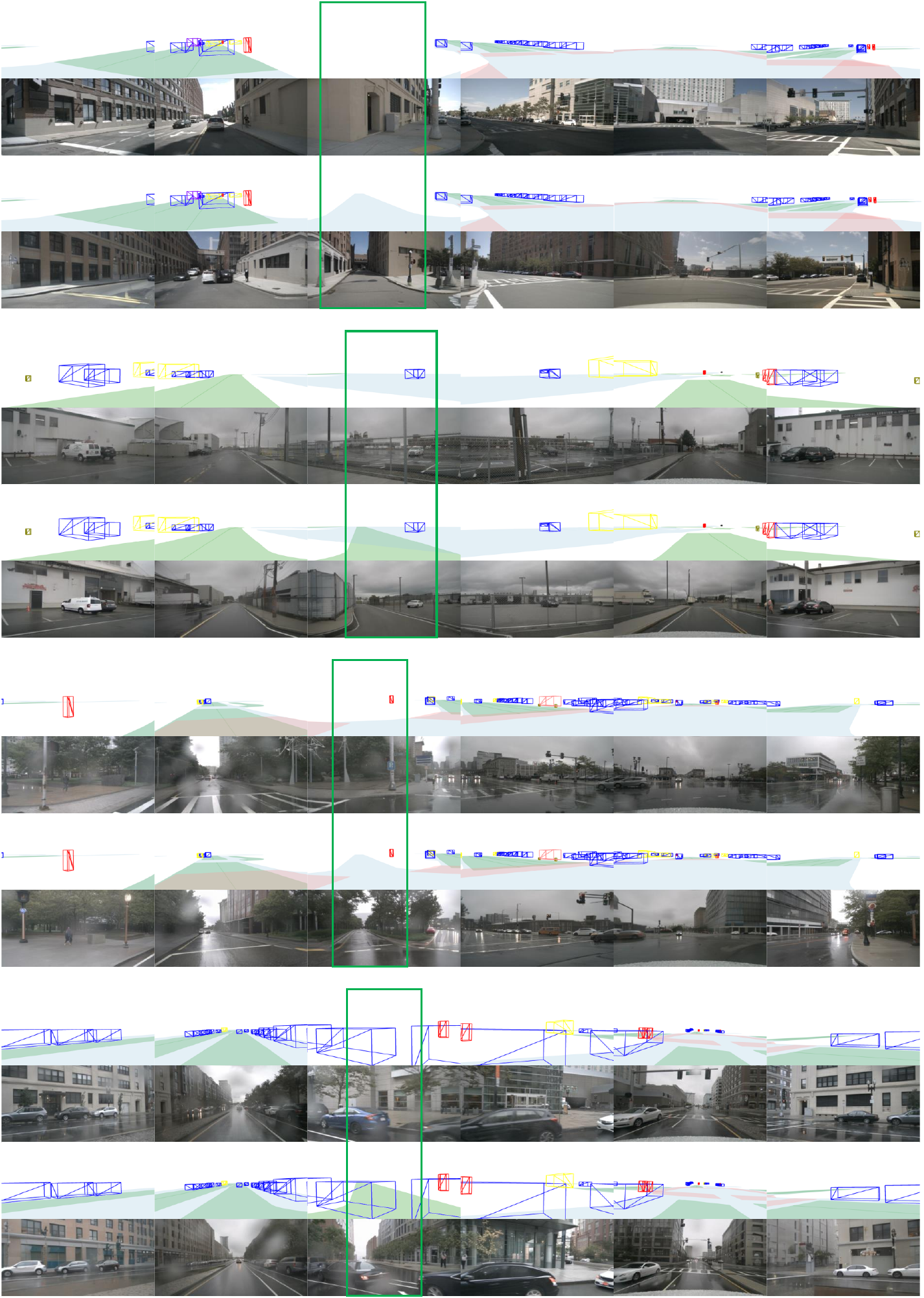}
	\caption{Additional examples of targeted road addition using our DrivePTS framework. The visualization format follows the same structure as above.}
	\label{fig.appendix-add-split-road-2}
\end{figure*}

Supplementary examples are provided to further demonstrate the superior capabilities of our proposed DrivePTS. 
The multi-view hierarchical descriptions significantly outperform the short captions from existing datasets regarding scene detail recovery, as illustrated in Fig.~\ref{fig.des_comparable_1} and \ref{fig.des_comparable_2}. 
These comprehensive descriptions enable a more nuanced and accurate reconstruction of complex driving scenarios.
Additionally, Fig.~\ref{fig.appendix-del-road}, \ref{fig.appendix-add-split-road-1}, and \ref{fig.appendix-add-split-road-2} demonstrate DrivePTS's ability to align with modified maps through successful road removal and addition across diverse environments. 
These results highlight the framework's adaptability to dynamic road configurations and demonstrate its potential for generating various road types to validate navigation tasks.

\subsection{Generalization to Video Generation}
We further evaluate the generalization capability of the proposed strategy and components on the video generation task.
Following the design of MagicDrive, we extend DrivePTS with spatio-temporal attention to model temporal dependencies.
For quantitative evaluation, we adopt the Fréchet Video Distance (FVD)~\cite{unterthiner2018towards} metric to measure temporal coherence and overall video quality.
The results are summarized below.

\begin{table}[H]
	\centering
	\caption{
		Comparison of temporal consistency for different video generation methods on the NuScenes dataset. 
		Lower FVD scores indicate better temporal coherence.
	}
	\label{Tab.video-fvd-horizontal}
	\resizebox{1.02\linewidth}{!}{
		\begin{tabular}{@{}l|cccccc@{}}
			\toprule
			Metric & MagicDrive & DriveDreamer & Panacea & Drive-WM & Ours (SD-2.1) & \textbf{Ours (SD-3.5)} \\
			\midrule
			FVD$\downarrow$ & 221 & 353 & 139 & 122 & 128 & \textbf{110} \\
			\bottomrule
		\end{tabular}
	}
\end{table}

As shown in Tab.~\ref{Tab.video-fvd-horizontal}, our temporal variant still achieves lower FVD scores than most existing baselines.
It is worth noting that Drive-WM\cite{wang2024driving} benefits from additional conditional inputs, including driving actions and control signals, enabling a future-aware world model with stronger temporal coherence.
Without introducing video-specific designs, our framework still shows strong adaptability, confirming that the proposed strategy and components can be effectively extended to temporal generation tasks.
Furthermore, when equipped with the more advanced SD3.5~\cite{esser2024scaling} backbone following the DiT paradigm, DrivePTS achieves the best FVD score, benefiting from DiT-based spatio-temporal modeling capacity.
These results highlight the generalizability and extensibility of DrivePTS.

\subsection{Limitation and Future Work}
Despite overall improvements, the fidelity of generated lane markings and traffic sign details remains suboptimal, indicating the need for more precise spatial and semantic control.
Future work will explore incorporating more advanced fine-grained constraints to further refine local structures and contextual consistency.
These enhancements are expected to facilitate the synthesis of more realistic and coherent driving scenes while supporting fine-grained perception tasks such as lane detection, traffic sign recognition, and signal understanding.

\clearpage
\begin{lstlisting}[basicstyle=\tiny\ttfamily, breaklines=true, frame=single, numbers=none]
	You are an image captioner specialized in autonomous driving scenes. 
	You will be provided with multiple images taken by the ego vehicle's 360-degree surround view cameras, with viewpoints at the "CAM_FRONT_LEFT", "CAM_FRONT", "CAM_FRONT_RIGHT", "CAM_BACK_RIGHT", "CAM_BACK", "CAM_BACK_LEFT" of the ego vehicle.
	
	Given multiple images from different camera viewpoints, you must output a valid JSON object with exactly the following structure:
	{
		"CAM_FRONT_LEFT": {
			"time": <time of day, choose from e.g. "daytime", "night", "evening", "dawn", "dusk">,
			"weather": <concise weather condition, choose from e.g. "clear", "sunny", "overcast", "cloudy", "foggy", "rainy", "drizzle", "snowy", "hazy", "stormy">,
			"road type": <brief description of the road surface, choose from "straight road", "split road", "left-turn lane", "right-turn lane", "cross-junction", "T-junction", "Y-junction", "roundabout", "merging road", "no road">,
			"surroundings": <brief description of the scene type, e.g. "urban intersection", "residential street", "multi-lane highway", "construction zone", "urban park area">,
			"static_and_dynamic_objects": <comma-separated list of visible and relevant static and dynamic entities, e.g. "cars, trucks, buses, bicycles, pedestrians, traffic lights, traffic signs, trees, buildings, cones, barriers">,
			"spatial relationships": <a rich and informative sentence that describes the relationship among the scene layout, interactions, or driving-relevant dynamics. Do not simply repeat values from the previous fields. Focus on spatial relationships, motion patterns, visual semantics, or notable features>},
		"CAM_FRONT": {
			// Same structure as above},
		"CAM_FRONT_RIGHT": {
			// Same structure as above},
		"CAM_BACK_RIGHT": {
			// Same structure as above},
		"CAM_BACK": {
			// Same structure as above},
		"CAM_BACK_LEFT": {
			// Same structure as above}
	}
	
	Constraints:
	- Note!! Each camera viewpoint's caption (including field names, punctuation, and values) must not exceed 110 tokens.
	- Output ONLY the JSON object--no additional text.
	- The images may be blurred due to rain or movement. If the text cannot be read clearly, please do not guess the content of the text blindly. So, for textual content, it is important to answer accurately.
	- For congestion: If there are other cars in front of ego vehicle at a close distance and there is a large flow of traffic in the same lane next to it, it can be considered congested.
	- Common static objects include traffic signs/signals, road infrastructure, road markings, road obstacles/barriers, parking facilities, traffic monitoring equipment, roadside facilities, roadside greenery, advertising/information boards, and public facilities.
	- Common dynamic objects include motor vehicles, non-motorized vehicles, pedestrians, emergency vehicles, construction equipment/vehicles, and public transportation vehicles.
	- If a camera viewpoint shows no clear road or driving scene, set "road type" to "no road" and adjust other fields accordingly.
	
	Example format:
	{
		"CAM_FRONT_LEFT": {
			"time": "daytime",
			"weather": "sunny, clear sky",
			"road type": "left-turn lane",
			"surroundings": "urban street scene",
			"static_and_dynamic_objects": "bus, car, fence, trees, building",
			"spatial relationships": "The image shows a street scene with a green fence, trees, and buildings. There's an orange bus on the road, and part of a car is visible in the foreground."},
		"CAM_FRONT": {
			"time": "daytime",
			"weather": "sunny, clear sky",
			"road type": "straight road",
			"surroundings": "urban intersection",
			"static_and_dynamic_objects": "traffic lights, cars, crosswalk, buildings",
			"spatial relationships": "Forward view shows an intersection with traffic lights overhead, vehicles waiting in lanes, and crosswalk markings visible on the road surface."}
		// ... continue for all other camera viewpoints
	}
\end{lstlisting}
\captionof{lstlisting}{Prompt template for VLM-based scene description generation.}
\label{lst:prompt}


\end{document}